\documentclass[conference]{IEEEtran}
\IEEEoverridecommandlockouts
\usepackage{cite}
\usepackage{amsmath,amssymb,amsfonts}
\usepackage{algorithmic}
\usepackage{graphicx}
\usepackage{textcomp}
\usepackage{xcolor}
\usepackage{multirow}
\usepackage{booktabs}
\usepackage{bm}
\usepackage{balance}
\usepackage{color}
\usepackage{bbm}
\usepackage{url}

\def\BibTeX{{\rm B\kern-.05em{\sc i\kern-.025em b}\kern-.08em
    T\kern-.1667em\lower.7ex\hbox{E}\kern-.125emX}}
\begin{document}

\title{Knowledge Enhanced Model for Live Video Comment Generation\\
}

\author{\IEEEauthorblockN{Jieting Chen}
\IEEEauthorblockA{\textit{School of Information} \\
\textit{Renmin University of China}\\
Beijing, China \\
jietingchen23@outlook.com}
\and
\IEEEauthorblockN{Junkai Ding}
\IEEEauthorblockA{\textit{Department of Statistics} \\
\textit{Columbia University}\\
New York, United States \\
jd3868@columbia.edu}
\and
\IEEEauthorblockN{Wenping Chen}
\IEEEauthorblockA{\textit{School of Information} \\
\textit{Renmin University of China}\\
Beijing, China \\
chenwenping@ruc.edu.cn}
\and
\IEEEauthorblockN{Qin Jin*\thanks{*Qin Jin is the corresponding author.}}
\IEEEauthorblockA{\textit{School of Information} \\
\textit{Renmin University of China}\\
Beijing, China \\
qjin@ruc.edu.cn}
}

\maketitle

\begin{abstract}
Live video commenting is popular on video media platforms, as it can create a chatting atmosphere and provide supplementary information for users while watching videos. Automatically generating live video comments can improve user experience and enable human-like generation for bot chatting. 
Existing works mostly focus on short video datasets while ignoring other important video types such as long videos like movies. 
In this work, we collect a new Movie Live Comments (MovieLC) dataset to support research on live video comment generation for long videos. 
We also propose a knowledge enhanced generation model inspired by the divergent and informative nature of live video comments. 
Our model adopts a pre-training encoder-decoder framework and incorporates external knowledge. 
Extensive experiments show that both objective metrics and human evaluation demonstrate the effectiveness of our proposed model. The MovieLC dataset and our code will be released.
\end{abstract}


\section{Introduction}
Many video platforms have the ``live video commenting" feature, which allows users to comment at any time while watching a video, and the comments instantly slide across the video screen, making users feel like watching together with others, even if they are watching alone. 
Live video commenting has become more and more popular among web users. 
The video platform Bilibili\footnote{\url{https://www.bilibili.com/}} reported that the number of live comments reached 10 billion in 2021 and there are more than 60 million live comment users on its platform\footnote{\url{https://new.qq.com/rain/a/20211129A0AKBB00}}. 
Live comments can also provide supplementary information that is related to but goes beyond the video content, which helps users understand the video, enjoy the video, and provoke discussion for more user interaction. Besides, live commenting technologies can be applied in bot chatting scenarios to support human-like user experience because of the flexible and divergent expressions. Therefore, automatic live video comment generation, as a meaningful and challenging task, has attracted increasing attention in the research community.

\begin{figure}
    \centering
    \includegraphics[width=0.9\linewidth]{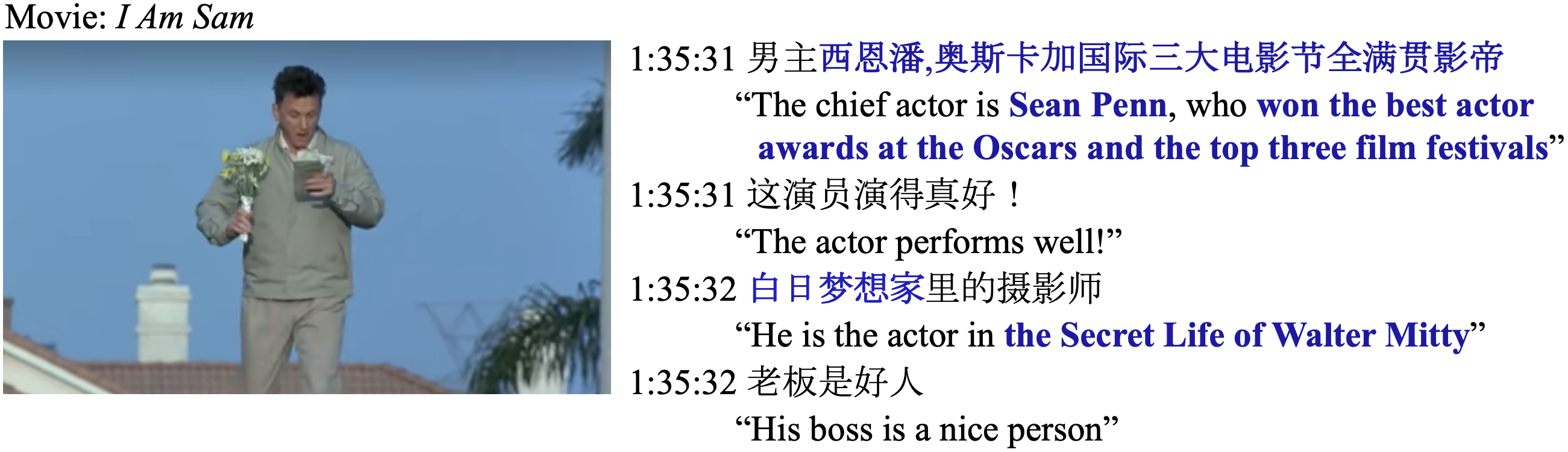}
    \caption{An example from MovieLC Dataset. The blue words associate with external knowledge that goes beyond the video content. Best viewed in color.}
    \label{fig:data_case}
\end{figure}

One might intuitively think that video comment generation is similar to video caption generation. However, we argue that video comments and video captions differ greatly in the data characteristics and pragmatics. Video captions are normally objective descriptions of the visual semantic content, while live video comments are very divergent but informative. 
 Existing live video comments datasets including Livebot \cite{ma2019livebot}, VideoIC \cite{wang2020videoic}, and the extensional Livebot \cite{wu2021cold}, are all collected from the same video platform Bilibili\footnote{\url{https://www.bilibili.com/}}, where most of the videos are user-generated short videos, and the live comments are often sloppy. They ignore other important types of videos such as long videos like movies, which may trigger more live comments and discussions due to the long video duration. 
In order to ensure sufficient exploration for the live comment generation task, we collect a new Movie Live Comments (MovieLC) dataset\footnote{https://github.com/AIM3-RUC/MovieLC}, which contains 1,406,219 live comments from 85 movies, totaling 175 hours. 
As shown in the examples in Fig.\ref{fig:data_case}, in addition to objective descriptions of the video content, live video comments may include divergent information associated with external knowledge, for example, the awards Sean Penn has won and other representative works in which he played the leading role, etc.

Most of the existing works on live comment generation follow the routines of video captioning approaches. The live video comments generation task is first defined in \cite{ma2019livebot}. The following works mainly focus on addressing the interactivity between modalities \cite{wang2020videoic, zeng2021plvcg} and the diversity of generation \cite{zhang2020dca, zeng2021plvcg}. 
These works typically ignore the informative and divergent features of live video comments. We argue that the generation model should introduce external knowledge beyond the visual and text context.
Therefore, we propose a \textbf{K}nowledge Enhanced Model for \textbf{L}ive \textbf{V}ideo \textbf{C}omment \textbf{G}eneration (\textbf{KLVCG}), which adopts a pre-training encoder-decoder framework inspired by \cite{luo2020univl}. Specifically, we design independent encoders for visual context, comment context, and external knowledge respectively. A cross encoder is built for interaction across modalities, and an auto-regressive decoder is constructed for the generation. We recall external knowledge from knowledge graphs and comments from other videos. Then we incorporate them via the masked language prediction task in the pre-training stage.

We standardize a more reasonable evaluation setting and conduct experiments on three datasets, including Livebot \cite{ma2019livebot}, VideoIC \cite{wang2020videoic} and MovieLC datasets. Our model significantly outperforms baselines on three datasets on both objective metrics and human evaluation. 

The main contributions of this work include: 1) We build MovieLC, a movie live comments dataset. The long movie videos with informative live comments can well complement existing datasets. 2) We propose the KLVCG model, which can generate higher-quality live video comments by effectively leveraging external knowledge. 3) Extensive experiments on three benchmark datasets validate that our model significantly outperforms other competing methods. 

\section{Related Works}
\noindent\textbf{Live video comment generation.}
Existing works on live comment generation typically follow the captioning routines. The task is firstly defined in \cite{ma2019livebot}, which releases a benchmark dataset called Livebot and proposes a model that employs transformers with multiple blocks. The following works on this task \cite{duan2020multimodal,wang2020videoic,zeng2021plvcg} mainly focus on the interactivity among different modalities. 
For example, a multi-modal transformer trained with a multi-task learning is proposed to enhance inter-modal and intra-modal interactions \cite{wang2020videoic}. 
Another line of works \cite{zhang2020dca,zeng2021plvcg} focus on the diversity of generation. 
Works on other aspects like cold start in generation \cite{wu2021cold} and timestamp prediction for generation \cite{wu2021knowing} are also proposed. All these works use user-generated short videos for validation.
Additionally, existing works ignore the divergent feature of comments, while we believe that introducing extra knowledge beyond visual content is important for better generation.

\noindent\textbf{Knowledge enhanced generation.}
Incorporating knowledge for text generation has shown promising performance on many tasks, such as Visual Question Answering (VQA) \cite{marino2021krisp, gao2022transform} and dialogue response generation \cite{zhou2022think, shen2022knowledge}. 
To the best of our knowledge, knowledge enhancement for video comment generation has not been well investigated. The key challenge is how to effectively fuse knowledge into model. Pre-training is a commonly used method, which incorporates the knowledge into the model by designing specific pre-training tasks like masked language modeling or entity/relation prediction \cite{zhang2019ernie, liu2020k, sun2020colake}. Some works implicitly incorporate the knowledge as feature  \cite{zhang2019ernie}. The knowledge can be embedded in a feature by the knowledge representation model such as TransE \cite{bordes2013translating}. The other works explicitly incorporate the knowledge as natural language tokens \cite{liu2020k,sun2020colake}, which is intuitive and effective. We choose the explicit way and leverage a pre-training task to introduce external knowledge.

\section{Datasets}
\noindent \textbf{MovieLC data collection.}
We select 85 famous movies and collect the live comments data from a public video platform\footnote{\url{https://v.qq.com/}}. We check the data manually to ensure that the live comments are correctly aligned with the visual content. 
In addition, we also acquire the movie scripts from the public website\footnote{\url{http://assrt.net}}, and the meta information including storylines, tags, and casts from Douban\footnote{\url{https://movie.douban.com/}} for future exploration. 
We split the data in MovieLC into training, validation, and testing sets by the ratio of 8:1:1 (more details can be found in the supplementary material). We will release the dataset and processed data for easy reproduction. To protect the copyrights of movies, we will only provide the IMDb urls following the same protocol that has been applied in other movie/TV shows datasets such as TRIPOD \cite{papalampidi2019movie} and VIOLIN \cite{liu2020violin}.

\begin{table}[t]
    \centering
    \small
    \caption{Statistics of three datasets.} 
    \begin{tabular}{llll}
    \toprule
\textbf{Dataset} & \textbf{Livebot\cite{ma2019livebot}} & \textbf{VideoIC\cite{wang2020videoic}} & \textbf{MovieLC} \\ \hline
\#Videos    &2,361  &4,951    &85 \\
\#Comments   &895,929   &5,330,393    &1,406,219  \\ 
Durations (h)  &114  &557    &175    \\ \hline 
Avg. Dur(s) &174   &405      &7412 \\
Avg. \#Comm    &380  &1,077   &16,544 \\
Avg. \#Char &9.08   &9.14      &10.37 \\
Avg. \#Word &5.42     &5.39    &6.53 \\ 
    \bottomrule
    \end{tabular}
    \label{tab:data_statistics}
\end{table}

\noindent \textbf{MovieLC data statistics.}
Our MovieLC dataset contains 1,406,219 live comments, aligned with 85 movies, spanning in total 175 hours. Table~\ref{tab:data_statistics} shows the statistics of three datasets. Different from previous datasets, MovieLC consists of long videos lasting 7412 seconds on average, and more comments with 16,544 comments per video on average. The lengths of characters/words are longer as well. It is because that the movie live comments are with more meaningful expressions and fewer buzzwords. As shown in Figure~\ref{fig:data_case}, the diversity and informativeness of movie live comments bring more challenges for the generation task.

\begin{table}[tb]
\caption{Notation table.}
\vspace{-8pt}
\label{tab:notation}
\centering
\small
\setlength{\tabcolsep}{1mm}{
\begin{tabular}{c|l}
\hline
Symbol & Definition  \\
\hline
$t$ & the target time step for comment generation \\ 
\hline
$l$ & the time range of the context window \\ 
\hline
$\bm{F}$=$\{\bm{f}_{i}\}$ & the context of frame sequence \\ 
\hline
$\bm{C}$=$\{\bm{c}_{i}\}$ & the context of comment sequences\\ 
\hline
$\bm{c}_{i}$=$\{\bm{w}_{ij}\}$ & the word sequence of comments at time step $i$ \\ 
\hline
$\bm{K}$=$\{\bm{k}_{i}\}$ & the word sequence of recalled knowledge \\ 
\hline
$\bm{Y}$=$\{\bm{y}_i\}$ & the word sequence of the target comment \\ 
\hline
$\bm{V}$=$\{\bm{v}_{i}\}$ & the feature sequence of context frames \\ 
\hline
$\mathbbm{M}$=$\{\bm{M}_{i}\}$ & the feature sequence of context comments \\ 
\hline
$\bm{W}$ & the output of the cross encoder \\
\hline
$\bm{D}$ & the output hidden states of the decoder \\
\hline
$\bm{P}$ & the output prediction of the model \\
\hline
$\mathcal{L}_{\mathrm{mask}}$ & the loss of pre-training task \\
\hline
$\mathcal{L}_{\mathrm{gen}}$ & the loss of fine-tuning task \\
\hline
\end{tabular}}
\vspace{-8pt}
\end{table}

\begin{figure*}[tb]
    \centering
    \includegraphics[width=0.9\linewidth]{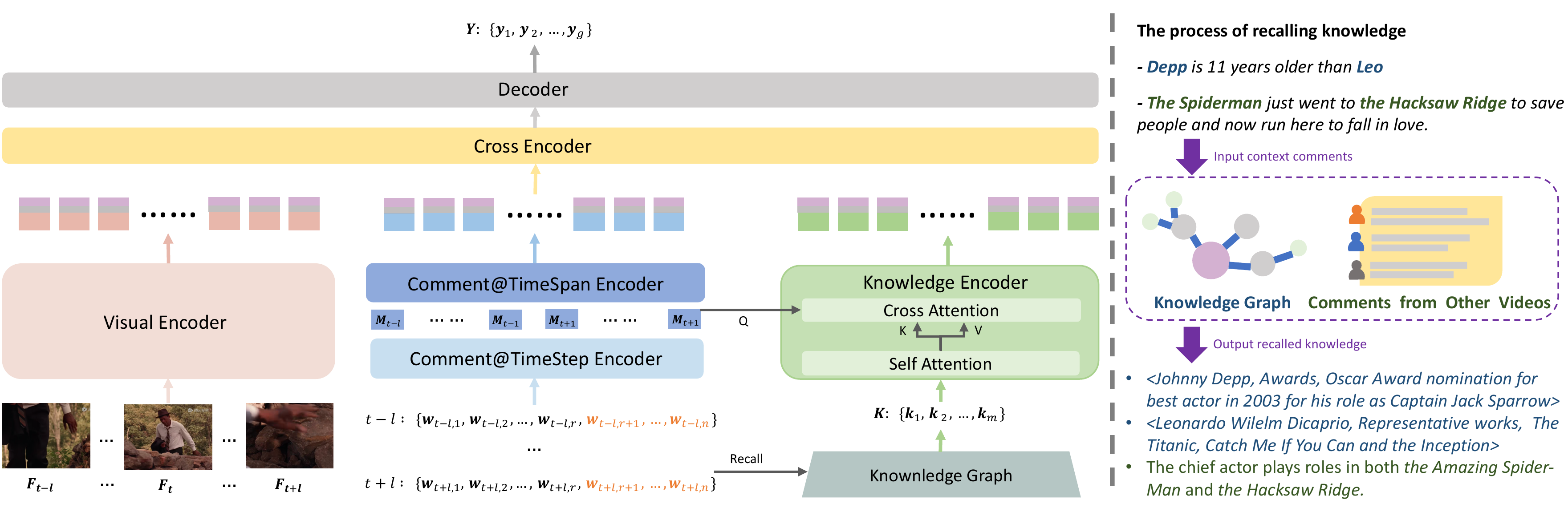}
    \vspace{-7pt}
    \caption{The left side shows the model structure of KLVCG. It incorporates recalled knowledge by a knowledge encoder. The comment context is augmented by related comments from other videos (in orange). The outputs of single modality encoders are added by token type embedding (in grey) and period type embedding (in purple). The right side shows the process of recalling knowledge from the knowledge graph (in blue) and comments from other videos (in green). Best viewed in color.}
    \label{fig:model}
\end{figure*}

\section{Methodology}
We propose a new framework for live video comment generation \textbf{KLVCG} that recalls external knowledge and integrates them to enhance the comment generation. At time step $t$, we consider the time range $[t$-$l, t$+$l]$ as its context window. Given the video context $\bm{F}$, comment context $\bm{C}$, and auxiliary external knowledge $\bm{K}$, KLVCG aims to generate a comment $\bm{Y}$=$\{\bm{y}_1, \bm{y}_2,..., \bm{y}_g\}$ at time step $t$, where $g$ denotes the maximum length of the generated words. Table~\ref{tab:notation} provides the exposition of symbols we used in the following part.



\subsection{Model architecture}
The left side of Fig.\ref{fig:model} illustrates the overall framework of our knowledge enhanced model for live video comment generation (KLVCG).
Following the encoder-decoder framework, KLVCG consists of independent modality encoders for video context, comment context, and external knowledge respectively, a cross encoder, and a decoder. All encoders are multi-layer transformers with the hidden size of $d$.

\noindent \textbf{Visual encoder.} 
Given the target time step $t$, we sample one frame per second from the context window $[t$-$l, t$+$l]$ and form the visual frame context $\bm{F}$=$\{\bm{f}_{t-l},\cdots,\bm{f}_t,\cdots,\bm{f}_{t+l}\}$ for the target time. The visual features are then extracted by a pre-trained ResNet101~\cite{he2016deep}, followed by a linear layer to project the feature to the $d$-dimensional visual context $\bm{V}$=$\{\bm{v}_{t-l},\cdots,\bm{v}_t,\cdots,\bm{v}_{t+l}\}\!\in\! \mathbb{R}^{(2l+1)\times{d}}$. Position embedding is added to preserve the temporal information. Finally, the contextualized encoding $\bm{V^{'}}$ is generated via a transformer encoder: $\bm{V^{'}}\in\mathbb{R}^{(2l+1)\times{d}}, \bm{V^{'}}=Transformer_v(\bm{V})$.

\noindent \textbf{Comment@TimeStep encoder.} 
At the target time step $t$, users may post the comment based on not only the visual content, but also the surrounding comments. 
As multiple comments normally appear at a time step, 
we concatenate the comments appearing at the same time step $i$ with SEP separator token and add a CLS token at the beginning. We then tokenize the sequence\footnote{\url{https://github.com/fxsjy/jieba}} 
into a word sequence $\bm{c}_i=\{\bm{w}_{i1}, \bm{w}_{i2},..., \bm{w}_{in}\}$, where ${n}$ denotes the maximum length of the sequence. After adding the position embedding, the Comment@TimeStep encoder 
produces the contextualized feature $\bm{{c_i}^{'}} \in \mathbb{R}^{n\times{d}}, \bm{c}_i^{'}=Transformer_m(\bm{c}_i)$.
After that, we use the contextualized CLS token to represent the comments at time step $i$. We simply denote it as $\bm{M}_i$.


\noindent \textbf{Comment@TimeSpan encoder.} 
The sequence of comments at each time step in the context window $\mathbbm{M}$=$\{\bm{M}_{t-l},...,\bm{M}_{t-1},$ $\bm{M}_{t+1},...,$ $\bm{M}_{t+l}\}$  with position embedding is fed to the Comment@TimeSpan encoder to produce contextualized comment context embedding. $\bm{{C}^{'}}\in \mathbb{R}^{2l\times{d}}, \bm{{C}^{'}}=Transformer_c(\mathbbm{M})$.

\noindent \textbf{Knowledge encoder.} We further incorporate external knowledge to enhance the informativeness and diversity of generated comments. 
We first recall relevant knowledge facts based on the comment context. We then concatenate and tokenize them into a word sequence  $\bm{K}$=$\{\bm{k}_1, \bm{k}_2,...,\bm{k}_m\}$, where $m$ denotes the maximum length of knowledge words. 
The knowledge encoder first performs self attention among knowledge words. Then, we use the Comment@TimeStep encoding $\bm{M}$=$\{\bm{M}_{t-l},...,\bm{M}_{t-1},\bm{M}_{t+1},...,\bm{M}_{t+l}\}$ as query to perform cross attention for modality interaction and denoising. The comment context is beneficial for attending to related information and ignoring the noise in the recalled knowledge. The Knowledge Encoder output $\bm{K^{'}} \in \mathbb{R}^{2l\times{d}}$ is produced by
\begin{align}
    \tilde{\bm{K}}&=SelfAttention_k(\bm{K},\bm{K},\bm{K}),\nonumber\\
    \bm{{K}^{'}}&=CrossAttention_k({\bm{M}},\bm{\tilde{K}},\bm{\tilde{K}}).
\end{align}

\noindent \textbf{Cross encoder.} 
The outputs of visual encoder, knowledge encoder, and Comment@TimeSpan encoder are then concatenated as the input sequence to the cross encoder for inter-modality interaction. We add a token type embedding to mark the modality of each token. Furthermore, we notice that viewing period is related to the user comment pattern. For example, users tend to express greetings at the beginning of the video. In the middle, live comments are more about the discussions of story and characters. While at the end, most comments are emotional expressions or farewell greetings. We quantify the viewing period into five stages and add a period type embedding. The output of the cross encoder is $\bm{W} \in \mathbb{R}^{(6l+1)\times{d}}, {\bm{W}}$=$Transformer_{cross}([\bm{V^{'}};\bm{C^{'}};\bm{K^{'}}])$.

\noindent \textbf{Decoder.} Based on the cross encoder output $\bm{W}$, a transformer decoder produces the hidden state $\bm{D} \in \mathbb{R}^{2l\times{d}}$ which is projected into prediction $\bm{P} \in \mathbb{R}^{g\times{vocab\_size}}$ by a linear layer, and then generates the comment $\bm{Y}$=$\{\bm{y}_1, \bm{y}_2,..., \bm{y}_g\}$. 
\begin{align}
    {\bm{D}}&=Transformer_{decoder}(\bm{W}),\nonumber\\
    {\bm{P}}&=Linear(\bm{D}).
\end{align}

\subsection{Knowledge enhancing}
As shown in the right side of Fig.\ref{fig:model}, KLVCG is enhanced by two types of external knowledge: the knowledge graph and the comments from other videos. \\
\noindent{\textbf{Enhancing by knowledge graph.}}
We choose Ownthink\footnote{\url{https://github.com/ownthink/KnowledgeGraphData}} as knowledge source for general datasets (Livebot and VideoIC) and FilmkG \cite{zhou2020kdconv} for the movie dataset (MovieLC). Ownthink is a public open-domain knowledge graph in Chinese with 140 million facts. FilmKG contains 8,090 entities and 98,661 facts for the film domain. All the facts are organized in the triplet format \textless entity1, relation, entity2\textgreater.
Then, we recall knowledge facts based on the keywords of context comments. We build the keyword set for each video by TF-IDF value. These words are used as searching entities to link the knowledge facts. Figure~\ref{fig:model} shows that our method is able to recall external knowledge facts like the awards and representative works of the actor, which is relevant to the content and helpful for enjoying the video. 


\noindent{\textbf{Enhancing by the comments from other videos.}}
The comments from other videos can also provide useful information. We recall comments about the same entities from other videos. Similar to the above strategy, we select noun words of context comments by TF-IDF value as key entities for each video. 
Then we concatenate retrieved comments with original context comments. The case in the right side of Fig.~\ref{fig:model} shows that the retrieved comment points out the semantic relation between \emph{the Spider-man} and \emph{the Hacksaw Ridge}. 

\subsection{Training strategy}
\noindent \textbf{Pre-training.} We incorporate the external knowledge via the masked language prediction task. We randomly mask 30\% words of the target comment $\bm{Y}$=$\{\bm{y}_1, \bm{y}_2,..., \bm{y}_g\}$. 
For these masked locations, 80\% of them are replaced by a [MASK] token, and 10\% are replaced by a random word. 
The pre-training objective is to predict the masked words $\bm{y}_\mathrm{m}$ based on the seen ones $\bm{y}_{\setminus \mathrm{m}}$. We use the cross-entropy loss as follows:
\begin{equation}
    \mathcal{L}_{\mathrm{mask}}=-\mathbb{E}_{\bm{y}_{\mathrm{m}}\sim \bm{Y}}\log p(\bm{y}_\mathrm{m}|\bm{y}_{\setminus \mathrm{m}},\bm{P}).
\end{equation}

\noindent \textbf{Fine-tuning.} 
The decoder generates the comments in an auto-regressive manner. We also use the cross-entropy loss in the fine-tuning stage:
\begin{equation}
    \mathcal{L}_{\mathrm{gen}}=-\mathbb{E}_{\bm{y}_i\sim \bm{Y}}\log p(\bm{y}_i|\bm{y}_{<i},\bm{P}).
\end{equation}
\section{Experiments}

\subsection{Experiment settings}\label{sec_settings}
\noindent\textbf{Evaluation setup.} We evaluate the comment generation performance in a ranking manner following previous works \cite{ma2019livebot, wu2020response}. The common practice is to ask the model to rank 100 candidate comments according to its generation probabilities. We average the cross-entropy losses of each token in the comment as the ranking score. The metrics for retrieval, including Recall@k (R@k), Mean Rank (MR), and Mean Reciprocal Rank (MRR) are reported. 
The candidate comments include 5 ground-truth comments (posted by viewers at the target TimeStep) and some negative comments chosen from the training set in 3 ways: 1) \textbf{Popular}: 20 most frequent comments. 2) \textbf{Plausible}: 20 comments with the highest TF-IDF cosine similarity to the context comments. 3) \textbf{Random}: the rest comments are picked randomly. Further details are put in the supplementary material.

\begin{table}[t]
\centering
\small
\caption{Comparison between UT (Unified Transformer) \cite{ma2019livebot}, MML-CG \cite{wang2020videoic}, and KLVCG. KLVCG+ represents results of our model pre-trained on the mixture of all three datasets.}
\vspace{-10pt}
\resizebox{\linewidth}{!}{
\begin{tabular}{llllll}
\\ \toprule
\multicolumn{1}{l|}{\textcolor{magenta}{\textbf{Livebot}}} & \textbf{R@1↑} & \textbf{R@5↑} & \textbf{R@10↑} & \textbf{MR↓} & \textbf{MRR↑} \\ \midrule
\multicolumn{1}{l|}{{UT}} & 9.65 & 35.48 & 53.62 & 14.98 & 0.230  \\
\multicolumn{1}{l|}{{MML-CG}} & 10.42 & 36.43 & 54.81 & 15.64 & 0.239 \\
\multicolumn{1}{l|}{{KLVCG}} & \textcolor{blue}{\textbf{13.49}} & \textcolor{blue}{\textbf{41.43}} & \textcolor{blue}{\textbf{59.31}} & \textcolor{blue}{\textbf{13.09}} & \textcolor{blue}{\textbf{0.276}}  \\ 
\multicolumn{1}{l|}{{KLVCG+}} & \textcolor{red}{\textbf{14.88}} & \textcolor{red}{\textbf{44.81}} & \textcolor{red}{\textbf{62.50}} & \textcolor{red}{\textbf{11.91}} & \textcolor{red}{\textbf{0.295}}  \\ 
\bottomrule
\\ \toprule
\multicolumn{1}{l|}{\textcolor{magenta}{\textbf{MovieLC}}} & \textbf{R@1↑} & \textbf{R@5↑} & \textbf{R@10↑} & \textbf{MR↓} & \textbf{MRR↑} \\ \midrule
\multicolumn{1}{l|}{{UT}} & 6.24 & 18.98 & \textcolor{red}{\textbf{31.98}} & \textcolor{blue}{\textbf{23.76}} & 0.147  \\
\multicolumn{1}{l|}{{MML-CG}}  & 6.25 & 17.81 & 30.64 & 24.70 & 0.143 \\
\multicolumn{1}{l|}{{KLVCG}} & \textcolor{blue}{\textbf{7.36}} & \textcolor{blue}{\textbf{19.26}} & 29.92 & 24.87 & \textcolor{blue}{\textbf{0.154}}  \\ 
\multicolumn{1}{l|}{{KLVCG+}} & \textcolor{red}{\textbf{8.01}} & \textcolor{red}{\textbf{20.51}} & \textcolor{blue}{\textbf{31.68}} & \textcolor{red}{\textbf{23.71}} & \textcolor{red}{\textbf{0.163}}  \\ 
\bottomrule
\\ \toprule
\multicolumn{1}{l|}{\textcolor{magenta}{\textbf{VideoIC}}} & 
\textbf{R@1↑} & \textbf{R@5↑} & \textbf{R@10↑} & \textbf{MR↓} & \textbf{MRR↑} \\ \midrule
\multicolumn{1}{l|}{{UT}} & 30.46 & 55.17 & 68.92 & 9.96 & 0.428  \\
\multicolumn{1}{l|}{{MML-CG}}  & 31.26 & 55.15 & 68.70 & 10.23 & 0.433 \\
\multicolumn{1}{l|}{{KLVCG}} & \textcolor{blue}{\textbf{34.08}} & \textcolor{blue}{\textbf{57.22}} & \textcolor{red}{\textbf{71.37}} & \textcolor{blue}{\textbf{9.51}} & \textcolor{blue}{\textbf{0.458}} \\ 
\multicolumn{1}{l|}{{KLVCG+}} & \textcolor{red}{\textbf{34.11}} & \textcolor{red}{\textbf{57.33}} & \textcolor{blue}{\textbf{71.32}} & \textcolor{red}{\textbf{9.43}} & \textcolor{red}{\textbf{0.459}}  \\ 
\bottomrule
\end{tabular}}
\label{tab:sota_comparison}
\end{table}

\noindent\textbf{Training setup.} We set the context window as $[t$-$5, t$+$5]$. The max generated word length is set as $g$=$20$, including [BOS] and [EOS] tokens. We adopt teacher-forcing for training and evaluation, and apply beam search for inference. The beam size is set to 5. For the encoder-decoder framework, the transformer hidden size $d$ is set to 384. The visual encoder, Comment@TimeStep encoder, and Comment@TimeSpan encoder all have 2 layers transformer blocks with 6 attention heads. The knowledge encoder consists of a self-attention block and a cross-attention block both with 6 attention heads. The cross encoder has 2 layers transformer blocks with 12 attention heads. The decoder has 1 layer transformer block with 12 attention heads. We run experiments on 4 1080Ti GPUs with batch size 128. The learning rate for both pre-training and fine-tuning is set as 0.0001 and decays in the same way as UniVL \cite{luo2020univl}. More details can be found in the supplementary material. 

\subsection{Main results}\label{sec_main_results}
We compare to two state-of-the-art baseline models: 1) \textbf{Unified Transformer}: The first live comment generation model that employs transformers with multiple blocks \cite{ma2019livebot}. 2) \textbf{MML-CG}: The model enhances modalities interaction by multi-modal transformer and multi-task learning \cite{wang2020videoic}.

In Table~\ref{tab:sota_comparison}, the 'KLVCG' row shows the results from our models that are pre-trained on the target dataset. Numbers in red mark the highest results, and numbers in blue are the second-best results. Our KLVCG model significantly outperforms baselines on all three datasets in almost all metrics.
On the user-generated short videos dataset Livebot, the Recall@1 is boosted from 9.65 to 13.49, the MRR is improved from 0.230 to 0.276. On the larger dataset VideoIC, the Recall@1 of our KLVCG model also increases from 30.46 to 34.08. On the long movie video dataset MovieLC, the performance metrics are lower than that on the other two datasets, which indicates more challenges in long video understanding and comment generation. However, the performance of KLVCG still improves by incorporating knowledge (from 6.24 to 8.01). When mixing all three datasets for pre-training, performances are further improved as shown in the 'KLVCG+' row. 

According to the results, knowledge enhancement is helpful for generating live video comments for both general short videos and long movie videos. Furthermore, the parameter size of our model (43M) is smaller than that of the MML-CG model (78M) and the Unified Transformer model (82M). It indicates that the performance improvement comes from our model design rather than more parameters.

\subsection{Robustness in sparse context scenarios}\label{sec_robust}
\begin{figure}[t]
    \centering
    \includegraphics[width=0.70\linewidth]{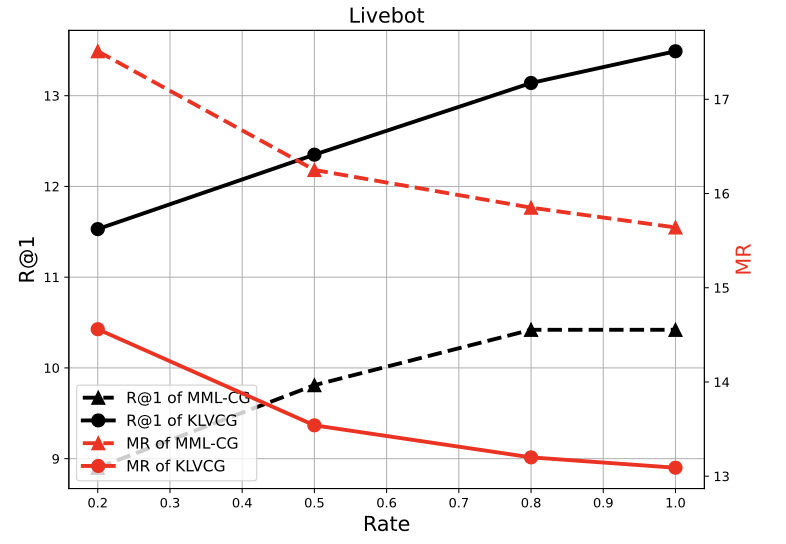}
    \vspace{-10pt}
    \caption{Performance under sparse context scenarios. The KLVCG model (solid lines) consistently outperforms the MML-CG baseline (dashed lines).}
    \label{fig:robustness}
\end{figure}

Fig.\ref{fig:robustness} shows the performance trend under different sparse context scenarios on the Livebot dataset (results on all three datasets are shown in the supplementary material). We reduce the context comment density to 80\%, 50\%, and 20\% respectively. Results show that even in sparse context scenarios, our model significantly outperforms MML-CG in a wide range of the context comment density. Experiment results demonstrate higher robustness of our model in sparse context scenarios, which is important for real applications.

\subsection{Ablation study}\label{sec_ablation}
We carry out the ablation study of KLVCG model framework on MovieLC and Livebot datasets. As reported in Table~\ref{tab:ablation_study}, we first ablate the performance of our framework trained from scratch without pre-training and other auxiliary information, which obviously outperforms the baselines as shown in the first row. Adding the pre-training stage boosts the performance (row 2). Both enhancing by knowledge graphs (row 3) and by comments from other videos (row 4) bring additional gain to the general video dataset Livebot and film domain dataset MovieLC. In addition, considering the period type is also beneficial to the generation process (row 5). In summary, all the designed components in our model contribute to the performance improvement.

\begin{table}[t]
\centering
\small
\caption{Ablation study. Basic: Our framework trained from scratch. MLM: Adding a pre-training stage with masked language prediction task. KG: Enhancing by the knowledge graph. AC: Enhancing by comments from other videos. Period: Adding the period embedding. }
\begin{tabular}{rrrrrr}
\toprule
\multicolumn{1}{l|}{\textcolor{blue}{\textbf{Livebot}}} & \textbf{R@1↑} & \textbf{R@5↑} & \textbf{R@10↑} & \textbf{MR↓} & \textbf{MRR↑} \\ \midrule
\multicolumn{1}{l|}{{Basic}} & 12.33 & 39.98 & 56.57 & 14.09 & 0.262  \\
\multicolumn{1}{l|}{{+MLM}} & 12.54 & 40.42 & 57.57 & 13.32 & 0.266  \\
\multicolumn{1}{l|}{{+KG}} & 12.90 & \textbf{41.61} & 58.90 & 13.18 & 0.272 \\
\multicolumn{1}{l|}{{+AC}} & 13.29 & 41.35 & 58.57 & 13.22 & 0.273 \\
\multicolumn{1}{l|}{{+Period}} & \textbf{13.49} & 41.43 & \textbf{59.31} & \textbf{13.09} & \textbf{0.276}  \\
\bottomrule
\\
\toprule
\multicolumn{1}{l|}{\textcolor{blue}{\textbf{MovieLC}}} & \textbf{R@1↑} & \textbf{R@5↑} & \textbf{R@10↑} & \textbf{MR↓} & \textbf{MRR↑} \\ \midrule
\multicolumn{1}{l|}{{Basic}} & 6.41 & 18.77 & 29.50 & 25.53 & 0.146  \\
\multicolumn{1}{l|}{{+MLM}} & 7.30 & 18.30 & 28.68 & 25.39 & 0.151  \\
\multicolumn{1}{l|}{{+KG}} & 7.30 & 19.03 & 29.32 & 25.00 & 0.153 \\
\multicolumn{1}{l|}{{+AC}} & 7.35 & 18.81 & 29.69 & \textbf{24.82} & 0.154 \\
\multicolumn{1}{l|}{{+Period}} & \textbf{7.36} & \textbf{19.26} & \textbf{29.92} & 24.87 & \textbf{0.154}  \\
\bottomrule
\end{tabular}
\label{tab:ablation_study}
\end{table}

\subsection{Human evaluation}\label{sec_human_eval}
We also carry out the human evaluation by recruiting 15 experienced live comments users. They are asked to choose the best comment out of candidates that are generated by Unified Transformer, MML-CG, and our model KLVCG. 
For the 900 testing samples, 44\% of the KLVCG generated comments are selected as the best, while only 15\% of the MML-CG generated comments and 11\% of the UT generated comments are selected as the best. Our KLVCG model clearly wins the preference of users. 

\begin{figure}
    \centering
    \includegraphics[width=0.9\linewidth]{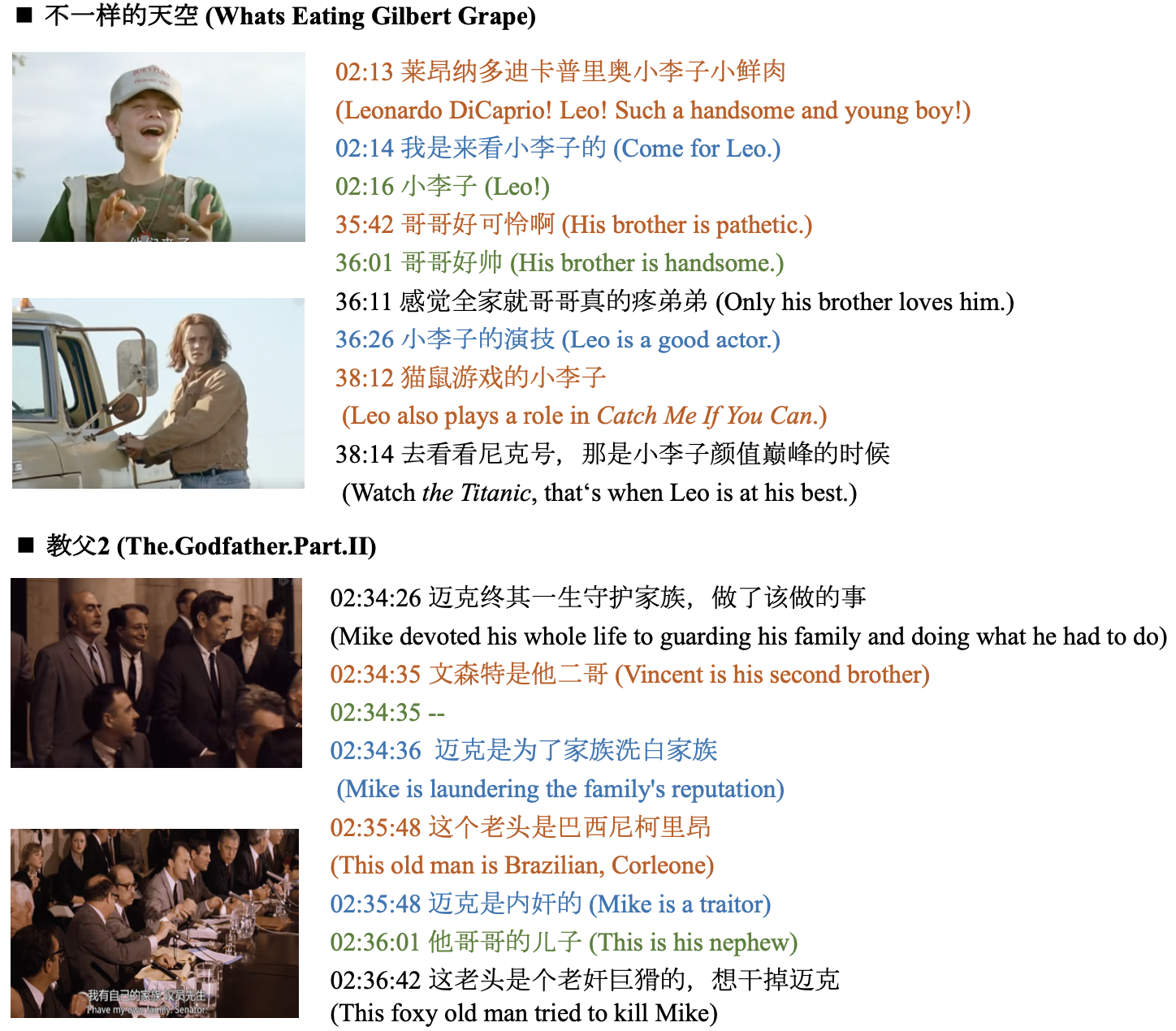}
    \caption{Generation case of KLVCG (in orange), MML-CG (in green), Unified Transformer (in blue), and the ground-truth (in black). Best viewed in color.}
    \label{fig:gen}
\end{figure}

\subsection{Case study}\label{sec_case_study}
Examples of generated comments on the MovieLC dataset are visualized in Fig.\ref{fig:gen}. Our model (in orange) can successfully introduce knowledge like the full name of the actor (\emph{Leonardo Dicaprio}) and his other masterpiece (\emph{Catch Me If You Can}), which do not appear in the original comment context. More cases can be found in the supplementary material.
Furthermore, we found that the KLVCG model can generate non-null comments for 97\% of the testing samples, while MML-CG and UT models can only reach 32\% and 62\%. It indicates that the knowledge enhancement helps the model to generate meaningful content in more scenarios. 
\section{Conclusion}
In this work, we build a Movie live comments (MovieLC) dataset to support research on live comments generation for long videos. We also propose a novel knowledge enhanced model (KLVCG), which leverages external knowledge to improve the diversity and informativeness of generated comments. Extensive experiments on three datasets demonstrate that our model significantly outperforms the competing approaches. The human evaluation and the case study further show that our model can generate more meaningful live comments. In the future work, we will further explore long video understanding and leveraging rich external knowledge.

\section{Acknowledgments}
This work was partially supported by National Natural Science Foundation of China (No. 62072462), National Key R\&D Program of China (No. 2020AAA0108600).

\bibliographystyle{IEEEtran}
\bibliography{IEEEabrv,mybibfile}

\begin{thebibliography}{10}
\providecommand{\url}[1]{#1}
\csname url@samestyle\endcsname
\providecommand{\newblock}{\relax}
\providecommand{\bibinfo}[2]{#2}
\providecommand{\BIBentrySTDinterwordspacing}{\spaceskip=0pt\relax}
\providecommand{\BIBentryALTinterwordstretchfactor}{4}
\providecommand{\BIBentryALTinterwordspacing}{\spaceskip=\fontdimen2\font plus
\BIBentryALTinterwordstretchfactor\fontdimen3\font minus
  \fontdimen4\font\relax}
\providecommand{\BIBforeignlanguage}[2]{{%
\expandafter\ifx\csname l@#1\endcsname\relax
\typeout{** WARNING: IEEEtran.bst: No hyphenation pattern has been}%
\typeout{** loaded for the language `#1'. Using the pattern for}%
\typeout{** the default language instead.}%
\else
\language=\csname l@#1\endcsname
\fi
#2}}
\providecommand{\BIBdecl}{\relax}
\BIBdecl

\bibitem{ma2019livebot}
S.~Ma, L.~Cui, D.~Dai, F.~Wei, and X.~Sun, ``Livebot: Generating live video
  comments based on visual and textual contexts,'' in \emph{Proceedings of the
  AAAI Conference on Artificial Intelligence}, vol.~33, no.~01, 2019, pp.
  6810--6817.

\bibitem{wang2020videoic}
W.~Wang, J.~Chen, and Q.~Jin, ``Videoic: A video interactive comments dataset
  and multimodal multitask learning for comments generation,'' in
  \emph{Proceedings of the 28th ACM International Conference on Multimedia},
  2020, pp. 2599--2607.

\bibitem{wu2021cold}
H.~Wu, F.~Piti{\'e}, and G.~Jones, ``Cold start problem for automated live
  video comments,'' in \emph{Proceedings of the Third Workshop on Multimodal
  Artificial Intelligence}, 2021, pp. 54--62.

\bibitem{zeng2021plvcg}
Z.~Zeng, N.~Gao, C.~Xue, and C.~Tu, ``Plvcg: A pretraining based model for live
  video comment generation,'' in \emph{Advances in Knowledge Discovery and Data
  Mining: 25th Pacific-Asia Conference, PAKDD 2021, Virtual Event, May 11--14,
  2021, Proceedings, Part II}.\hskip 1em plus 0.5em minus 0.4em\relax Springer,
  2021, pp. 690--702.

\bibitem{zhang2020dca}
Z.~Zhang, Z.~Yin, S.~Ren, X.~Li, and S.~Li, ``Dca: Diversified co-attention
  towards informative live video commenting,'' in \emph{Natural Language
  Processing and Chinese Computing: 9th CCF International Conference, NLPCC
  2020, Zhengzhou, China, October 14--18, 2020, Proceedings, Part II 9}.\hskip
  1em plus 0.5em minus 0.4em\relax Springer, 2020, pp. 3--15.

\bibitem{luo2020univl}
H.~Luo, L.~Ji, B.~Shi, H.~Huang, N.~Duan, T.~Li, J.~Li, T.~Bharti, and M.~Zhou,
  ``Univl: A unified video and language pre-training model for multimodal
  understanding and generation,'' \emph{arXiv preprint arXiv:2002.06353}, 2020.

\bibitem{duan2020multimodal}
C.~Duan, L.~Cui, S.~Ma, F.~Wei, C.~Zhu, and T.~Zhao, ``Multimodal matching
  transformer for live commenting,'' \emph{arXiv preprint arXiv:2002.02649},
  2020.

\bibitem{wu2021knowing}
H.~Wu, G.~J.~F. Jones, and F.~Pitie, ``Knowing where and what to write in
  automated live video comments: A unified multi-task approach,'' in
  \emph{Proceedings of the 2021 International Conference on Multimodal
  Interaction}, 2021, pp. 619--627.

\bibitem{marino2021krisp}
K.~Marino, X.~Chen, D.~Parikh, A.~Gupta, and M.~Rohrbach, ``Krisp: Integrating
  implicit and symbolic knowledge for open-domain knowledge-based vqa,'' in
  \emph{Proceedings of the IEEE/CVF Conference on Computer Vision and Pattern
  Recognition}, 2021, pp. 14\,111--14\,121.

\bibitem{gao2022transform}
F.~Gao, Q.~Ping, G.~Thattai, A.~Reganti, Y.~N. Wu, and P.~Natarajan,
  ``Transform-retrieve-generate: Natural language-centric outside-knowledge
  visual question answering,'' in \emph{Proceedings of the IEEE/CVF Conference
  on Computer Vision and Pattern Recognition}, 2022, pp. 5067--5077.

\bibitem{zhou2022think}
P.~Zhou, K.~Gopalakrishnan, B.~Hedayatnia, S.~Kim, J.~Pujara, X.~Ren, Y.~Liu,
  and D.~Hakkani-Tur, ``Think before you speak: Explicitly generating implicit
  commonsense knowledge for response generation,'' in \emph{Proceedings of the
  60th Annual Meeting of the Association for Computational Linguistics (Volume
  1: Long Papers)}, 2022, pp. 1237--1252.

\bibitem{shen2022knowledge}
S.~Shen, V.~P{\'e}rez-Rosas, C.~Welch, S.~Poria, and R.~Mihalcea, ``Knowledge
  enhanced reflection generation for counseling dialogues,'' in
  \emph{Proceedings of the 60th Annual Meeting of the Association for
  Computational Linguistics (Volume 1: Long Papers)}, 2022, pp. 3096--3107.

\bibitem{zhang2019ernie}
Z.~Zhang, X.~Han, Z.~Liu, X.~Jiang, M.~Sun, and Q.~Liu, ``Ernie: Enhanced
  language representation with informative entities,'' \emph{arXiv preprint
  arXiv:1905.07129}, 2019.

\bibitem{liu2020k}
W.~Liu, P.~Zhou, Z.~Zhao, Z.~Wang, Q.~Ju, H.~Deng, and P.~Wang, ``K-bert:
  Enabling language representation with knowledge graph,'' in \emph{Proceedings
  of the AAAI Conference on Artificial Intelligence}, vol.~34, no.~03, 2020,
  pp. 2901--2908.

\bibitem{sun2020colake}
T.~Sun, Y.~Shao, X.~Qiu, Q.~Guo, Y.~Hu, X.~Huang, and Z.~Zhang, ``Colake:
  Contextualized language and knowledge embedding,'' \emph{arXiv preprint
  arXiv:2010.00309}, 2020.

\bibitem{bordes2013translating}
A.~Bordes, N.~Usunier, A.~Garcia-Duran, J.~Weston, and O.~Yakhnenko,
  ``Translating embeddings for modeling multi-relational data,'' \emph{Advances
  in neural information processing systems}, vol.~26, 2013.

\bibitem{papalampidi2019movie}
P.~Papalampidi, F.~Keller, and M.~Lapata, ``Movie plot analysis via turning
  point identification,'' \emph{arXiv:1908.10328}, 2019.

\bibitem{liu2020violin}
J.~Liu, W.~Chen, Y.~Cheng, Z.~Gan, L.~Yu, Y.~Yang, and J.~Liu, ``Violin: A
  large-scale dataset for video-and-language inference,'' in \emph{Proceedings
  of the IEEE/CVF Conference on Computer Vision and Pattern Recognition}, 2020,
  pp. 10\,900--10\,910.

\bibitem{he2016deep}
K.~He, X.~Zhang, S.~Ren, and J.~Sun, ``Deep residual learning for image
  recognition,'' in \emph{Proceedings of the IEEE conference on computer vision
  and pattern recognition}, 2016, pp. 770--778.

\bibitem{zhou2020kdconv}
H.~Zhou, C.~Zheng, K.~Huang, M.~Huang, and X.~Zhu, ``Kdconv: A chinese
  multi-domain dialogue dataset towards multi-turn knowledge-driven
  conversation,'' \emph{arXiv preprint arXiv:2004.04100}, 2020.

\bibitem{wu2020response}
H.~Wu, G.~J. Jones, and F.~Pitie, ``Response to livebot: Generating live video
  comments based on visual and textual contexts,'' \emph{arXiv preprint
  arXiv:2006.03022}, 2020.

\end{thebibliography}

\end{document}